\documentclass{article}

\usepackage{authblk}
\usepackage[utf8]{inputenc} 
\usepackage[T1]{fontenc}    
\usepackage{hyperref}       
\usepackage{url}            
\usepackage{booktabs}       
\usepackage{amsfonts}       
\usepackage{nicefrac}       
\usepackage{microtype}      
\usepackage{xcolor}         
\usepackage{}
\usepackage{graphicx}       
\usepackage{multirow}       
\usepackage{array}          
\usepackage{threeparttable} 
\usepackage{adjustbox}      
\usepackage{tabularx}       
\usepackage{subcaption}     
\usepackage{tikz}           
\usepackage{amsmath, amssymb} 
\usepackage{pgfplots}       
\usepackage{pgfplotstable}  
\pgfplotsset{compat=1.18}
\usetikzlibrary{shapes.geometric, arrows, positioning, fit, backgrounds, calc, decorations.pathreplacing}
\usepackage{pdflscape}      
\usepackage{afterpage}      
\usepackage{mathtools}      
\usepackage{amsthm}         
\usepackage{natbib}
\usepackage[left=1.0in, right=1.0in, top=1.0in, bottom=1.0in]{geometry}
\title{Interpretable Fine-Gray Deep Survival Model for Competing Risks: Predicting Post-Discharge Foot Complications for Diabetic Patients in Ontario}
\usepackage{authblk}

\author[1]{Dhanesh Ramachandram\thanks{Corresponding author: \texttt{dhanesh.ramachandram@vectorinstitute.ai}}}
\author[2]{Anne Loefler}
\author[2,3]{Surain Roberts}
\author[2]{Amol Verma}
\author[1]{Maia Norman}
\author[2]{Fahad Razak}
\author[4]{Conrad Pow}
\author[4]{Charles de Mestral}

\affil[1]{Vector Institute \quad}
\affil[2]{GEMINI \quad}
\affil[3]{University of Toronto \quad}
\affil[4]{Diabetes Action Canada}

\date{}
\begin{document}

\maketitle

\begin{abstract}
  
    Model interpretability is crucial for establishing AI safety and clinician trust in medical applications for example, in survival modelling with competing risks. Recent deep learning models have attained very good predictive performance but their limited transparency, being black-box models,  hinders their integration into clinical practice. To address this gap, we propose an intrinsically interpretable survival model called CRISPNAM-FG. Leveraging the structure of Neural Additive Models (NAMs) with separate projection vectors for each risk, our approach predicts the Cumulative Incidence Function using the Fine-Gray formulation, achieving high predictive power with intrinsically transparent and auditable predictions. We validated the model on several benchmark datasets and applied our model to predict future foot complications in diabetic patients across 29 Ontario hospitals (2016-2023). Our method achieves competitive performance compared to other deep survival models while providing transparency through shape functions and feature importance plots.

\end{abstract}

\section{Introduction}

While traditional survival analysis focused on predicting risk for a single event of interest, many real-world problems, particularly in the healthcare domain, involve more than one ``competing'' events of interest. For example, patients who are at risk of death due to cancer may die following other unrelated complications. In this case, death due to cancer is the primary risk while death due to other causes is the competing risk that must be taken into account when predicting the primary risk. Traditionally, competing risks have been addressed using statistical methods such as the Fine-Gray model~\citep{fine1999proportional} and cause-specific hazard models, which rely on strong parametric assumptions about the relationship between covariates and risk. While these classical approaches offer interpretability, they are limited in their ability to capture complex, non-linear patterns in the data. Recent advances in deep learning have demonstrated improved predictive performance for competing risks scenarios~\citep{lee2018deephit}, but these gains come at the cost of interpretability, resulting in black-box models that are difficult for clinicians to trust and validate in practice.
In this work, we present an intrinsically interpretable survival model called CRISPNAM-FG (Competing Risks Survival Prediction using Neural Additive Models: Fine Gray) that handles competing risks using the Fine-Gray formulation while maintaining the predictive power of deep learning and the interpretability of classical statistical model

The remainder of this paper proceeds as follows. We begin by reviewing related work on deep neural network-based competing risks survival models, outlining key research gaps. Next, Section~\ref{methodology} provides the mathematical background and details our interpretable model architecture and its design principles. Section~\ref{experiments} describes the benchmark and GEMINI datasets, along with the procedures for model training, optimization, and evaluation. Finally, Section~\ref{Results} presents our results and discussion.
\section{Related Work}
\label{relatedwork}

Conventional survival analysis approaches, including the Kaplan-Meier estimator~\citep{goel2010understanding} and Cox proportional hazards model~\citep{breslow1975analysis}, are inappropriate for competing risk scenarios because they erroneously handle alternative events as censored observations, resulting in inflated estimates of event probabilities.

Several deep learning methodologies have been developed to handle competing risks in survival analysis. DeepHit~\citep{lee2018deephit} presents a joint modeling framework that employs a shared representation network coupled with cause-specific sub-networks to model the joint distribution of event time and type. The model optimization combines negative log-likelihood with ranking loss to ensure concordance between predicted and observed outcomes. Neural Fine Gray~\citep{jeanselme2023neural} extends the classical Fine-Gray sub-distribution model through neural networks, enabling non-linear covariate relationships while preserving direct cumulative incidence function estimation capabilities.

Despite these methodological advances, existing deep survival approaches for competing risks~\citep{jeanselme2023neural,lee2018deephit} lacks inherent interpretability, particularly at the feature level. This opacity makes it challenging to understand individual feature contributions to risk predictions across different competing events. Researchers thus have relied upon post-hoc explanations such as LIME~\citep{ribeiro2016should} and SHAP~\citep{lundberg2017unified}. Another example is SurvNAM~\citep{utkin2022survnam}, a post-hoc explanation method that approximates the predictions of a pre-trained black-box survival model (such as Random Survival Forest) by fitting a GAM-extended Cox PH model using NAMs as surrogate learners. Post hoc explanations, generally suffer from low fidelity, leading to unfaithful or misleading explanations and can be easily fooled~\citep{slack2020fooling}. 

Recently, glass box models or intrinsically interpretable models for survival analysis have been proposed. CoxNAM~\citep{xu2023coxnam}  embeds Neural Additive Models~\citep{agarwal2021neural} (NAMs) directly within the Cox framework, enabling end-to-end learning of nonlinear feature effects from survival data. Each feature is processed by its own neural network, and the results are summed to obtain the hazard function value. Interpretability is provided through shape functions and feature importance plots, however, this model is limited to single risk predictions. Inspired by CoxNAM, an intrinsically interpretable model called CRISPNAM~\citep{ramachandram2025crisp} was introduced to support competing risks through the \emph{cause-specific} framework. This model uses separate projection functions for each risk-feature pair to jointly learn all cause-specific hazards in a single framework, enabling features to have different effects across competing events while avoiding the need to treat competing events as censoring or fit separate models. However, cause-specific approaches impose independence assumptions and can produce biased cumulative incidence estimates when competing events are dependent. To address these limitations, we propose CRISPNAM-FG, which adapts the CRISPNAM architecture to implement the Fine-Gray formulation~\citep{fine1999proportional}, enabling direct modeling of cumulative incidence functions while preserving the interpretability and computational efficiency of the original framework.

\section{Methodology}
\label{methodology}

\subsection{Competing Risks in Survival Analysis}

There are two main methodologies for estimation in competing risks survival analysis, where subjects can experience one of K mutually exclusive event types, with $E \in \{0, 1, 2, \ldots, K\}$ where $E = 0$ denotes censoring.

\subsubsection{The Cause-Specific Hazard Method}
Proposed by~\citet{prentice1978analysis}, this method models competing events separately, treating each outcome as a distinct hazard function. Subjects experiencing competing events are removed from the risk set (treated as censored). For each cause $k$, the cause-specific hazard $\lambda_k(t|\mathbf{x})$ represents the instantaneous rate of event type $k$ at time $t$:

\begin{equation}
\lambda_k(t|\mathbf{x}) = \lim_{\Delta t \to 0} \frac{P(t \leq T < t + \Delta t, E = k \mid T \geq t, \mathbf{x})}{\Delta t}
\end{equation}

where $\mathbf{x}$ represents covariates, $T$ the event time, and the conditioning on $T \geq t$ indicates the subject has survived event-free until time $t$.

\subsubsection{Fine-Gray Subdistribution Method}
Introduced by~\citet{fine1999proportional}, this method directly models the subdistribution hazard, maintaining subjects who experience competing events in the risk set. The subdistribution risk set for event $k$ at time $t$ is:

\begin{equation}
\mathcal{R}_k^{\text{sub}}(t) = \{j: T_j \geq t\} \cup \{j: T_j < t \text{ and } E_j \neq k \text{ and } E_j \neq 0\}
\end{equation}

This includes both subjects event-free at time $t$ and those who experienced a competing event before time $t$.

The subdistribution hazard is defined as:
\begin{equation}
\lambda_k^{\text{sub}}(t|\mathbf{x}) = \lim_{\Delta t \to 0} \frac{P(t \leq T < t + \Delta t, E = k \mid T \geq t \text{ or } (T < t \text{ and } E \neq k), \mathbf{x})}{\Delta t}
\end{equation}

The relationship between cause-specific and subdistribution hazards is:
\begin{equation}
\lambda_k^{\text{sub}}(t|\mathbf{x}) = \lambda_k(t|\mathbf{x}) \cdot r_k(t|\mathbf{x})
\end{equation}

where $r_k(t|\mathbf{x}) = P(T \geq t \mid T \geq t \text{ or } E \neq k) / P(T \geq t)$ is the reduction factor.

\subsection{Model Architecture}

\begin{figure}
    \centering
    \begin{adjustbox}{max width=\textwidth}
\usetikzlibrary{positioning, calc, shapes.multipart}

\begin{tikzpicture}[
  scale=0.75, transform shape,
  feature/.style={rectangle, draw=blue!70, fill=blue!10, rounded corners=3pt, minimum width=2.5cm, minimum height=1cm, font=\large},
  featurenet/.style={rectangle, draw=green!70, fill=green!10, rounded corners=5pt, minimum width=3.5cm, minimum height=1cm, font=\large},
  representation/.style={rectangle, draw=orange!70, fill=orange!10, rounded corners=3pt, minimum width=1.7cm, minimum height=1cm, font=\large},
  riskproj/.style={rectangle, draw=black!70, fill=white, rounded corners=3pt, minimum width=3.5cm, minimum height=0.6cm, font=\large},
  sumcircle/.style={circle, draw=black, fill=gray!10, minimum size=0.9cm, font=\large},
  riskagg/.style={rectangle, draw=gray!80, fill=gray!10, rounded corners=5pt, minimum width=3.2cm, minimum height=1cm, font=\large, font=\bfseries},
  riskscore/.style={rectangle, draw=blue!70, fill=blue!10, rounded corners=3pt, minimum width=2.3cm, minimum height=1cm, font=\large},
  riskscore2/.style={rectangle, draw=red!70, fill=red!10, rounded corners=3pt, minimum width=2.3cm, minimum height=1cm, font=\large},
  fgsection/.style={rectangle, draw=purple!70, fill=purple!10, rounded corners=5pt, minimum width=22cm, minimum height=4.5cm, font=\large},
  baseline1/.style={rectangle, draw=blue!70, fill=blue!20, rounded corners=3pt, minimum width=3cm, minimum height=1cm, font=\large},
  baseline2/.style={rectangle, draw=red!70, fill=red!20, rounded corners=3pt, minimum width=3cm, minimum height=1cm, font=\large},
  subhazard1/.style={rectangle, draw=blue!70, fill=blue!10, rounded corners=3pt, minimum width=3.5cm, minimum height=1cm, font=\large},
  subhazard2/.style={rectangle, draw=red!70, fill=red!10, rounded corners=3pt, minimum width=3.5cm, minimum height=1cm, font=\large},
  cif1/.style={rectangle, draw=blue!70, fill=blue!20, rounded corners=3pt, minimum width=3cm, minimum height=1cm, font=\large},
  cif2/.style={rectangle, draw=red!70, fill=red!20, rounded corners=3pt, minimum width=3cm, minimum height=1cm, font=\large},
  formula/.style={rectangle, draw=gray!50, fill=white, rounded corners=3pt, font=\large},
  fgformula/.style={rectangle, draw=purple!50, fill=purple!5, rounded corners=3pt, font=\small},
  >=stealth,
]

\node[feature] (x1) at (0, 0) {Feature $x_1$};
\node[feature, below=2.0cm of x1] (x2) {Feature $x_2$};
\node[below=0.5cm of x2] (dots1) {$\vdots$};
\node[feature, below=1.1cm of dots1] (xp) {Feature $x_p$};

\node[featurenet, right=1.5cm of x1] (f1) {FeatureNet $f_1(x_1)$};
\node[featurenet, right=1.5cm of x2] (f2) {FeatureNet $f_2(x_2)$};
\node[right=1.5cm of dots1] (dots2) {$\vdots$};
\node[featurenet, right=1.5cm of xp] (fp) {FeatureNet $f_p(x_p)$};

\node[representation, right=1.5cm of f1] (h1) {$\mathbf{h}_1$};
\node[representation, right=1.5cm of f2] (h2) {$\mathbf{h}_2$};
\node[right=1.5cm of dots2] (dots3) {$\vdots$};
\node[representation, right=1.5cm of fp] (hp) {$\mathbf{h}_p$};

\node[riskproj, above right=0.1cm and 1.0cm of h1] (g11) {$g_{1,1}(\mathbf{h}_1)$};
\node[riskproj, above right=0.1cm and 1.0cm of h2] (g21) {$g_{2,1}(\mathbf{h}_2)$};
\node[riskproj, above right=0.1cm and 1.0cm of hp] (gp1) {$g_{p,1}(\mathbf{h}_p)$};
\node[font=\bfseries\Large, above=0.1cm of g11.north] {Projections};

\node[riskproj, below right=0.1cm and 1.0cm of h1] (g12) {$g_{1,2}(\mathbf{h}_1)$};
\node[riskproj, below right=0.1cm and 1.0cm of h2] (g22) {$g_{2,2}(\mathbf{h}_2)$};
\node[riskproj, below right=0.1cm and 1.0cm of hp] (gp2) {$g_{p,2}(\mathbf{h}_p)$};

\node[sumcircle, right=2.8cm of g12] (sum1) {+};
\node[sumcircle, right=2.8cm of g22] (sum2) {+};

\node[riskagg, right=1.2cm of sum1] (agg1) {Additive Aggregator};
\node[riskagg, right=1.2cm of sum2] (agg2) {Additive Aggregator};

\node[riskscore, right=1.2cm of agg1] (eta1) {$\eta_1(\mathbf{x})$};
\node[riskscore2, right=1.2cm of agg2] (eta2) {$\eta_2(\mathbf{x})$};
\node[font=\bfseries\Large, above=0.1cm of eta1.north] {Subdistribution log-hazard};

\node[formula, below=3.8cm of eta2,xshift=-2cm] (formula) {$\lambda_k^{FG}(t|\mathbf{x}) = \lambda_{0k}^{FG}(t) \cdot \exp(\eta_k(\mathbf{x}))$};

\node[fgsection, below=2.5cm of formula.south] (fgbox) {};
\node[font=\bfseries\Large, below=0.1cm of fgbox.south] {Fine-Gray: Baseline CIF Estimation \& Direct CIF Prediction};

\node[baseline1, left=7cm of fgbox.center, yshift=1cm] (F01) {$F_{01}(t)$};
\node[baseline2, left=7cm of fgbox.center, yshift=-1cm] (F02) {$F_{02}(t)$};
\node[font=\footnotesize, below=0.1cm of F01] {Baseline CIF 1};
\node[font=\footnotesize, below=0.1cm of F02] {Baseline CIF 2};

\node[subhazard1, left=1.5cm of fgbox.center, yshift=1cm] (lambdafg1) {$\lambda_1^{FG}(t|\mathbf{x})$};
\node[subhazard2, left=1.5cm of fgbox.center, yshift=-1cm] (lambdafg2) {$\lambda_2^{FG}(t|\mathbf{x})$};

\node[fgformula, right=1.5cm of lambdafg1] (cif1formula) {$F_1^{FG}(t|\mathbf{x}) = 1 - (1 - F_{01}(t))^{\exp(\eta_1(\mathbf{x}))}$};
\node[fgformula, right=1.5cm of lambdafg2] (cif2formula) {$F_2^{FG}(t|\mathbf{x}) = 1 - (1 - F_{02}(t))^{\exp(\eta_2(\mathbf{x}))}$};

\node[cif1, right=7cm of fgbox.center, yshift=1cm] (F1) {$F_1^{FG}(t|\mathbf{x})$};
\node[cif2, right=7cm of fgbox.center, yshift=-1cm] (F2) {$F_2^{FG}(t|\mathbf{x})$};
\node[font=\footnotesize, below=0.1cm of F1] {CIF 1};
\node[font=\footnotesize, below=0.1cm of F2] {CIF 2};

\foreach \i/\f/\h in {x1/f1/h1, x2/f2/h2, xp/fp/hp} {
  \draw[->] (\i) -- (\f);
  \draw[->] (\f) -- (\h);
}

\foreach \h/\gA/\gB in {h1/g11/g12, h2/g21/g22, hp/gp1/gp2} {
  \draw[->, blue!70, thick] (\h) -- (\gA.west);
  \draw[->, red!70, thick] (\h) -- (\gB.west);
}

\foreach \g in {g11, g21, gp1} {
  \draw[->, blue!70, thick] (\g.east) -- (sum1);
}
\draw[->, blue!70, thick] (sum1) -- (agg1);
\draw[->, blue!70, thick] (agg1) -- (eta1);

\foreach \g in {g12, g22, gp2} {
  \draw[->, red!70, thick] (\g.east) -- (sum2);
}
\draw[->, red!70, thick] (sum2) -- (agg2);
\draw[->, red!70, thick] (agg2) -- (eta2);

\draw[->, dashed] (eta1.east) -- ++(+2, 0) |- (formula);
\draw[->, dashed] (eta2.east) -- ++(+2, 0) |- (formula);

\draw[->, blue!70, thick] (F01) -- (lambdafg1);
\draw[->, red!70, thick] (F02) -- (lambdafg2);

\draw[->, blue!70, thick, dashed] (F01.east) to[out=0,in=180] (cif1formula.west);
\draw[->, red!70, thick, dashed] (F02.east) to[out=0,in=180] (cif2formula.west);

\draw[->, blue!70, thick] (lambdafg1) -- (cif1formula);
\draw[->, red!70, thick] (lambdafg2) -- (cif2formula);

\draw[->, blue!70, thick] (cif1formula) -- (F1);
\draw[->, red!70, thick] (cif2formula) -- (F2);

\draw[->, dashed] (formula.south) -- ($(formula.south)!0.5!(fgbox.north)$) -- (fgbox.north);

\node[font=\bfseries\footnotesize, purple, above=0.2cm of formula] {Fine-Gray Subdistribution Hazards};

\node[font=\footnotesize, purple, left=0.5cm of fgbox.west, text width=3cm, align=center] {Risk set includes subjects with competing events};

\end{tikzpicture}
    \end{adjustbox}
    \caption{CRISPNAM-FG Architecture Diagram}
    \label{fig:arch-diag}
\end{figure}
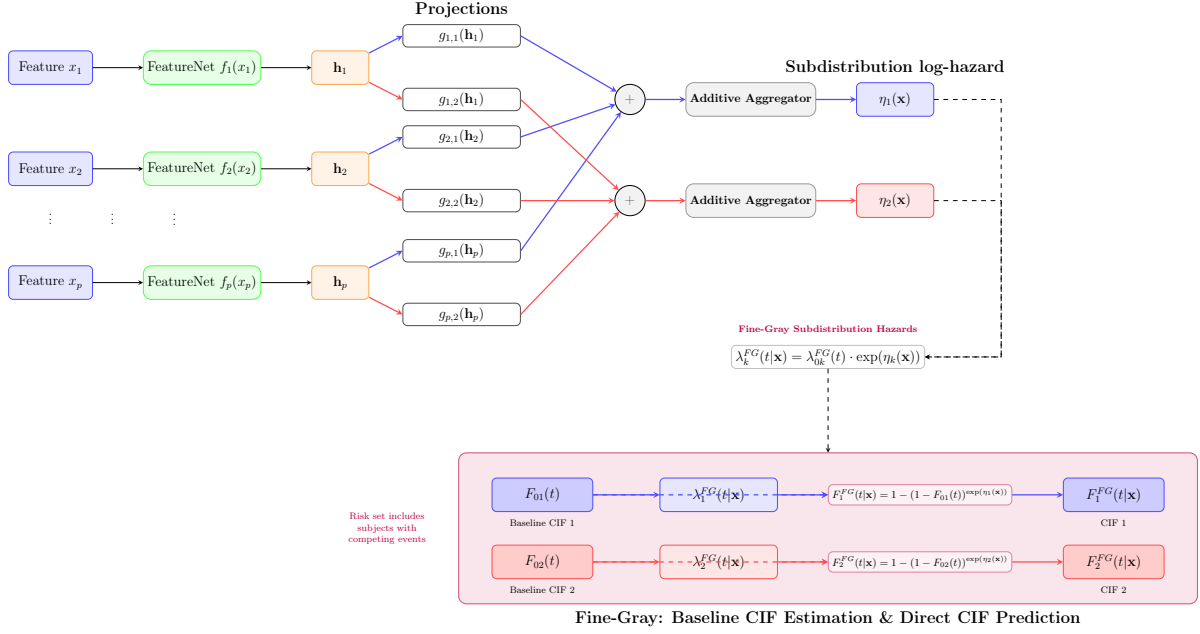

The CRISPNAM-FG model architecture is illustrated in Figure~\ref{fig:arch-diag}.

\subsubsection{Neural Additive Model (FeatureNet)}

In line with the Neural Additive Model framework, each input feature $x_i$ is processed by its own dedicated neural network $f_i(\cdot)$, referred to here as a \emph{FeatureNet}. These feature-specific sub-networks are designed to learn the non-linear contribution of each individual feature to the overall risk score, while preserving interpretability by isolating feature effects. 

\begin{equation}
\mathbf{h}_i = f_i(x_i) \in \mathbb{R}^d
\end{equation}

where $d$ is the dimension of the hidden representation. Each FeatureNet is a fully-connected feedforward neural network with $L$ layers, taking the scalar input $x_i$ and producing a hidden representation $\mathbf{h}_i \in \mathbb{R}^d$. The activations are computed recursively using the hyperbolic tangent function:

\begin{equation}
\mathbf{z}^{(l)}_i = \tanh(W^{(l)}_i \mathbf{z}^{(l-1)}_i + b^{(l)}_i), \quad \text{for } l = 1, \ldots, L,
\end{equation}

where $\mathbf{z}^{(0)}_i = x_i$, and $\mathbf{h}_i = \mathbf{z}^{(L)}_i$ denotes the output of the final layer. Each layer $l$ has weights $W^{(l)}_i \in \mathbb{R}^{d_l \times d_{l-1}}$ and biases $b^{(l)}_i \in \mathbb{R}^{d_l}$, with $d_0 = 1$ and $d_L = d$.

In this implementation, Dropout is used with rate $p_{\text{dropout}}$ after each hidden layer, Feature dropout with rate $p_{\text{feature}}$ during training to increase robustness and an optional batch normalization layer after each linear transformation to stabilize learning especially for deeper FeatureNets. 

\subsection{Risk-Specific Projections}

For each feature $i$ and competing risk $k$, a separate linear projection transforms the feature representation to its contribution to the log-hazard ratio. To address scale ambiguities across different competing risks and ensure fair comparison of feature contributions, we constrain the projection vectors to have unit L2 norm.

The risk-specific projection is defined as:
\begin{equation}
g_{i,k}(\mathbf{h}_i) = \tilde{\mathbf{w}}_{i,k}^T \mathbf{h}_i
\end{equation}
where $\tilde{\mathbf{w}}_{i,k}$ is the L2-normalized projection vector:
\begin{equation}
\tilde{\mathbf{w}}_{i,k} = \frac{\mathbf{w}_{i,k}}{\|\mathbf{w}_{i,k}\|_2 + \epsilon}
\end{equation}
with $\mathbf{w}_{i,k} \in \mathbb{R}^d$ being the learnable weight vector for projection $i \in \{1, 2, \ldots, p\}$ and risk $k \in \{1, 2, \ldots, K\}$, and $\mathbf{h}_i \in \mathbb{R}^d$ being the feature representation.

This normalization constraint ensures that $\|\tilde{\mathbf{w}}_{i,k}\|_2 = 1$ for all feature-risk pairs, which constraints all projection vectors to operate on the same scale and enabling direct comparison of feature importance across different competing risks.  

\subsubsection{Additive Risk Aggregation}

The log-hazard ratio for risk $k$ given input features $\mathbf{x} = [x_1, x_2, \ldots, x_p]$ is computed as the sum of individual feature contributions:
\begin{equation}
\eta_k(\mathbf{x}) = \sum_{i=1}^{p} g_{i,k}(f_i(x_i))
\end{equation}
This preserves the additive nature of the model while allowing for complex non-linear feature effects.

\subsubsection{Fine-Gray Subdistribution Hazards}

The Fine-Gray approach directly models the subdistribution hazard, which has a more direct relationship with the cumulative incidence function. For a subject with covariates $\mathbf{x}$, the subdistribution hazard for event type $k$ is parameterized as:
\begin{equation}
\lambda_k^{\text{sub}}(t|\mathbf{x}) = \lambda_{0k}^{\text{sub}}(t)\exp(\eta_k(\mathbf{x}))
\end{equation}

where $\lambda_{0k}^{\text{sub}}(t)$ is the baseline subdistribution hazard.

\subsubsection{Fine-Gray Partial Likelihood Loss}

The Fine-Gray negative log partial likelihood differs significantly from the cause-specific approach in its risk set construction. For event type $k$, the loss function is:

\begin{equation}
\mathcal{L}_k^{\text{FG}} = -\sum_{i: E_i = k} \left[ \eta_k(\mathbf{x}_i) - 
\log\left(\sum_{j \in \mathcal{R}_k^{\text{sub}}(T_i)} w_j(T_i) \exp(\eta_k(\mathbf{x}_j))\right) \right]
\end{equation}

where $w_j(T_i)$ are inverse probability of censoring weights:
\begin{equation}
w_j(T_i) = \frac{I(C_j \geq T_i)}{\hat{G}(T_i)}
\end{equation}

with $\hat{G}(t)$ being the Kaplan-Meier estimate of the censoring distribution.

For numerical stability and to handle class imbalance, we implement a weighted version:
\begin{equation}
\mathcal{L}^{\text{FG}} = \sum_{k=1}^{K} \omega_k \mathcal{L}_k^{\text{FG}} + \gamma \|\Theta\|^2_2
\end{equation}

where $\omega_k = n/(K \cdot n_k)$ for class balancing, and $\gamma \geq 0$ is the L2 regularization coefficient controlling the strength of weight decay.

\subsubsection{Cumulative Incidence Function Estimation}

Following Fine-Gray methodology, the cumulative incidence function for event $k$ is:
\begin{equation}
F_k(t|\mathbf{x}) = 1 - \exp\left(-\int_0^t \lambda_k^{\text{sub}}(s|\mathbf{x}) ds\right)
\end{equation}

In practice, using the Breslow-type estimator:
\begin{equation}
\hat{F}_k(t|\mathbf{x}) = 1 - \left[1 - \hat{F}_{0k}(t)\right]^{\exp(\eta_k(\mathbf{x}))}
\end{equation}

where $\hat{F}_{0k}(t)$ is the baseline cumulative incidence function estimated from:
\begin{equation}
\hat{F}_{0k}(t) = \sum_{i: T_i \leq t, E_i = k} \frac{d_i}{\sum_{j \in \mathcal{R}_k^{\text{sub}}(T_i)} \exp(\eta_k(\mathbf{x}_j))}
\end{equation}

with $d_i$ being the number of events at time $T_i$.

\subsection{Interpretability Mechanisms}

To faciliate interpretability of the model, \emph{shape functions plots}  can be generated to visualize the contribution of each feature to the prediction. Specifically, for each feature $i$ and risk $k$, we can compute a shape function that describes how the feature affects the log-hazard ratio.

\begin{equation}
s_{i,k}(x_i) = g_{i,k}(f_i(x_i))
\end{equation}

The importance of feature $i$ for risk $k$ is quantified by the mean absolute value of its contribution across the dataset.
\begin{equation}
\mathcal{I}_{i,k} = \frac{1}{N} \sum_{j=1}^{N} \left| s_{i,k}(x_{ij}) \right|
\end{equation}

Note that the feature importance metric $\mathcal{I}_{i,k}$ measures only the \textbf{magnitude} of a feature's influence on the competing risk, not its \textbf{direction}. Since $\mathcal{I}_{i,k}$ is computed using absolute values of the shape functions, a high importance score indicates that feature $i$ has a strong effect on risk $k$, but does not reveal whether this effect is protective (decreasing risk) or harmful (increasing risk).  To understand the actual risk relationship, one must examine the shape function plots $s_{i,k}(x_i)$ directly, where positive values indicate risk-increasing effects and negative values indicate risk-reducing effects. This separation of effect magnitude (captured by $\mathcal{I}_{i,k}$) from effect direction (shown in $s_{i,k}$) enables practitioners to first identify which features matter most for risk stratification, and then understand how these influential features actually modulate patient risk.

\section{Experiments}
\label{experiments}

\subsection{Benchmark Datasets}
\label{sec:datasets}
We compared the performance of our  proposed model against two other black box models for competing risks prediction namely Neural Fine-Gray~\citep{jeanselme2023neural} and DEEPHIT~\citep{lee2018deephit} using the following three real-world medical datasets:

\paragraph{Primary Biliary Cholangitis (PBC).} The PBC dataset originates from a randomized controlled trial conducted at the Mayo Clinic between 1974 and 1984, involving 312 patients diagnosed with primary biliary cholangitis. The study aimed to evaluate the efficacy of D-penicillamine in treating the disease. Each patient record includes 25 covariates encompassing demographic, clinical, and laboratory measurements. The primary endpoint was mortality while on the transplant waiting list, with liver transplantation considered a competing risk \citep{therneau2000cox}.

\paragraph{Framingham Heart Study.} Initiated in 1948, the Framingham Heart Study is a longitudinal cohort study designed to investigate cardiovascular disease (CVD) risk factors. For this analysis, data from 4,434 male participants were utilized, each with 18 baseline covariates collected over a 20-year follow-up period. The study focuses on modelling the risk of developing CVD, treating mortality from non-CVD causes as a competing event~\cite{kannel1979diabetes}.

\paragraph{SUPPORT2 Dataset.} The SUPPORT2 dataset originates from the Study to Understand Prognoses and Preferences for Outcomes and Risks of Treatments (SUPPORT2), a comprehensive investigation conducted across five U.S. medical centers between 1989 and 1994. This dataset encompasses records of 9,105 critically ill hospitalized adults, each characterized by 42 variables detailing demographic information, physiological measurements, and disease severity indicators. The study was executed in two phases: Phase I (1989–1991) was a prospective observational study aimed at assessing the care and decision-making processes for seriously ill patients, while Phase II (1992–1994) implemented an intervention to enhance end-of-life care. The primary objective was to develop and validate prognostic models estimating 2- and 6-month survival probabilities, thereby facilitating improved clinical decision-making and patient-physician communication regarding treatment preferences and outcomes \citep{support1995controlled}.

\subsection{GEMINI Dataset}

One of this paper's key contribution is applying the interpretable CRISPNAM-FG survival model to predict the probability of future foot complications in diabetes patients discharged from General Internal Medicine services at 29 Ontario hospitals.
Among the many complications that diabetic patients experience, diabetic foot complications represent a major concern, with individuals facing more than a 30\% lifetime risk of developing such complications and a 40\% recurrence rate within one year following wound healing~\citep{waibel2024current}.
Research demonstrates that annual foot screening can reduce both treatment costs and amputation rates~\citep{allen2023annual}. \citet{staniszewska2024effectiveness} reported that amputation rates can be reduced by 17\% to 96\% across multiple studies. While primary care data provides the optimal source for patient monitoring, hospitalization records from General Internal Medicine departments serve as strong predictors for identifying diabetic patients at risk of developing foot complications~\citep{boyko2006prediction}. 

In this work, a foot complication was defined as the first hospitalization for a foot ulcer, infection, gangrene, or (toe,foot,leg) amputation. Candidate predictors were specified a priori and on the basis of literature~\citep{dewi2020foot} describing the risk of foot complications in patients with diabetes. The dataset used for this work was obtained from GEMINI~\citep{gemini2024}, a non-profit organization that systematically collects administrative and clinical data from hospital information systems across Ontario. During the study period (April 2015-March 2023), GEMINI comprised 32 hospitals representing $50-60\%$ of medical and intensive care unit beds across Ontario, Canada. The network incorporates administrative data from the Canadian Institute for Health Information's Discharge Abstract Database and National Ambulatory Care Reporting System, alongside clinical data extracted from hospital-specific electronic medical records and mapped by clinical experts.

Of the 32 GEMINI hospitals, three hospitals were excluded: one lacked hemoglobin A1C data (a key model predictor), and two specialized cancer and rehabilitation facilities did not align with our target population. The cohort of index hospitalizations was comprised of all adult patients with diabetes discharged from general internal medicine between April 2016 and March 2023. Patients with diabetes were identified based on the presence of any ICD-10-CA code for diabetes (E10 – E14) in emergency department or inpatient records. Only patients discharged home were included to ensure eligibility for ambulatory foot screening. 

Patients with recorded age $>105$ years and those without a valid Ontario Health Insurance Plan (OHIP) number were excluded (the latter being required for linkage of different encounters to the same patient). Any encounters with a current foot complication (defined in Risk Calculator Outcome section below) were also excluded from the cohort of index admissions. Given that GEMINI data provided a 1-year lookback period, hospitalizations preceded by a foot complication in the prior year were excluded. Encounters from time periods with systematically missing laboratory data and those discharged during the final 30 days of the study period were removed since GEMINI data could not capture potential outcome encounters for these patients. When multiple eligible encounters existed for a single patient, the first hospitalization over the study period was selected as the index hospitalization.

\subsubsection{GEMINI Dataset Cohort Creation and Cohort Characteristics}

Table~\ref{tab:patient_characteristics} shows patient characteristics in the final cohort selected for future foot complication risk prediction dataset.

\begin{figure}[htbp]
    \centering
    \includegraphics[width=0.75\linewidth]{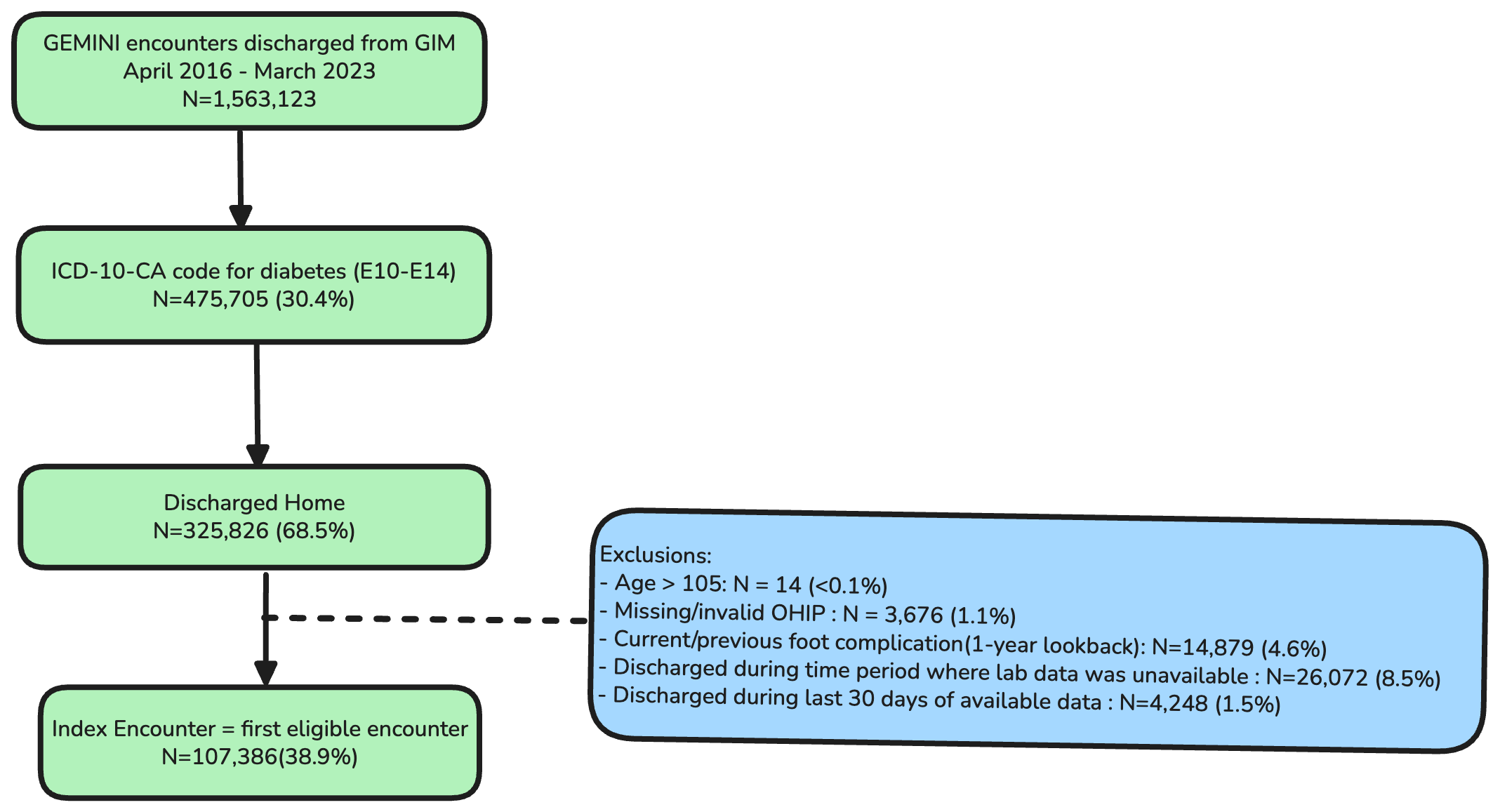}
    \caption{Cohort Generation Flow}
    \label{fig:cohort-gen}
\end{figure}

\begin{table}[htbp]
\centering
\caption{Patient characteristics (N = 107,836) for cohort in the GEMINI Dataset}
\label{tab:patient_characteristics}
\begin{tabular}{lc}
\toprule
\textbf{Characteristic} & \textbf{Value} \\
\midrule
Age (years), Median [Q1, Q3] & 72.0 [61.0, 81.0] \\
Sex = Female & 49,784 (46.2\%) \\
Elective admission & 1,989 (1.8\%) \\
\addlinespace
\textbf{Charlson Comorbidity Index (at admission)} & \\
\quad 0 & 34,294 (31.8\%) \\
\quad 1 & 42,957 (39.8\%) \\
\quad 2+ & 30,585 (28.4\%) \\
\addlinespace
Peripheral artery disease & 1,304 (1.2\%) \\
Coronary artery disease & 13,028 (12.1\%) \\
Stroke & 5,029 (4.7\%) \\
Congestive heart failure & 14,456 (13.4\%) \\
Hypertension & 48,104 (44.6\%) \\
COPD & 6,577 (6.1\%) \\
Chronic kidney disease & 4,405 (4.1\%) \\
Malignancy & 12,238 (11.3\%) \\
Mental illness & 7,483 (6.9\%) \\
Homelessness & 378 (0.4\%) \\
\addlinespace
\textbf{HbA1c (\%)} & \\
\quad Median [Q1, Q3] & 7.4 [6.4, 9.4] \\
\quad Missing: N (\%) & 64,828 (60.1\%) \\
\addlinespace
\textbf{Creatinine (umol/L)} & \\
\quad Median [Q1, Q3] & 84.0 [65.0, 115.3] \\
\quad Missing: N (\%) & 1,571 (1.5\%) \\
\addlinespace
\textbf{Albumin (g/L)} & \\
\quad Median [Q1, Q3] & 34.0 [29.4, 38.0] \\
\quad Missing: N (\%) & 33,566 (31.1\%) \\
\bottomrule
\end{tabular}
\end{table}

\subsubsection{Outcome and Feature Selection}

Table~\ref{tab:features} summarizes the study methods and features that were selected to train the risk prediction model.

\begin{table}[htbp]
\centering
\caption{Study Methods Summary}
\small
\label{tab:features}
\begin{tabular}{lr}
\hline
\textbf{Component} & \textbf{Description} \\
\hline
\multicolumn{2}{l}{\textbf{Primary Outcome}} \\
Definition & Time to first foot complication \\
Complications & Ulceration, infection, gangrene, amputation \\
Identification & Validated ICD-10-CA diagnosis and procedure codes \\
Setting & 29 GEMINI hospitals \\
Timing & Index hospitalization to first complication admission \\
Censoring & Last available date for patients without complications \\
Competing risk & In-hospital death \\
\hline
\multicolumn{2}{l}{\textbf{Predictors}} \\
Demographics & Age (continuous years), sex (female vs. non-female) \\
Clinical & Admission urgency (elective vs. urgent), homelessness \\
Laboratory & HbA1C, creatinine, albumin (last recorded values) \\
Comorbidities & Peripheral artery disease, coronary artery disease, \\
& stroke, congestive heart failure, hypertension, \\
& chronic obstructive pulmonary disease, chronic \\
& kidney disease, malignancy, mental illness \\
Coding & Binary variables from ICD-10-CA codes (index hospitalization) \\
Exclusions & Hospital-level factors excluded \\
\hline
\end{tabular}
\end{table}

Three laboratory variables contained missing values: albumin, creatinine, and HbA1C. Laboratory test ordering depends on physicians' clinical judgment regarding the potential utility of abnormal results and the availability of recent outpatient values, creating informative missingness patterns. To address this issue, we implemented the missing indicator method (MIM) whereby missing laboratory values were replaced with -1, and corresponding binary missingness indicators were created for each affected variable.

\subsection{Experimental Setup} 

 Our CRISPNAM-FG model is implemented using PyTorch. For benchmark datasets, we implemented nested K-fold cross validation with hyperparameter tuning. However, for the large GEMINI dataset, we held out 10\% of each dataset for hyperparameter optimization using Optuna~\citep{optuna} (learning rate, $L_2$ regularization strength, dropout rates, network architecture (1-3 hidden layers with 8-128 units), and batch normalization) using validation loss as the objective. We then performed 5-fold cross validation to obtain robust and unbiased estimate of the model's performance on the remaining 90\% of each datasets. Continuous features were normalized using standard scaling ($\mu=0, \sigma=1$) and categorical features were one-hot encoded. For the benchmark datasets, missing categorical values were imputed using mode imputation. For continuous variables, mean imputation was used. For missing lab values from the GEMINI dataset, we implemented the missing indicator method. Training employed the \textit{AdamW}~\citep{adamWpaper} optimizer to minimize the negative log-likelihood loss with a batch size of 256 and early stopping (patience=10) to prevent over-fitting.

Model performance was assessed using complementary discrimination and calibration metrics~\citep{park2021review} summarized in Table~\ref{tab:metrics}.
\begin{table}[ht!]
\caption{Performance metrics for model discrimination and calibration assessment}
\small
\centering
\begin{tabular}{lp{5cm}p{6cm}}
\hline
\textbf{Metric} & \textbf{Description} & \textbf{Interpretation} \\
\hline
TD-AUC & Discrimination ability: quantifies how well the model ranks subjects by risk & Values range from 0.5 (no better than chance) to 1.0 (perfect discrimination) \\
\hline
TD-CI & Discriminative ability that accounts for temporal changes in model performance & Values range from 0.5 to 1.0, with higher values indicating better discriminative ability; considers that model performance can evolve as time progresses \\
\hline
Brier Score & Accuracy of probabilistic predictions; penalizes errors in both discrimination and calibration & Lower values indicate better performance \\
\hline
\end{tabular}
\label{tab:metrics}
\end{table}

\section{Results and Discussion}
\label{Results}

Benchmark performance results are presented in Table~\ref{Tab:results}. We compared our CRISPNAM-FG model performance against two other models namely the Neural Fine-Gray~\citep{jeanselme2023neural} (NFG) and DEEPHIT~\citep{lee2018deephit}. The comparative analysis reveals distinct performance patterns between CRISPNAM-FG, DEEPHIT, and NFG models across the three competing risks datasets. 

\begin{table}[htb!]
\caption{Comparative 5-fold performance metrics for various models across multiple competing risks datasets.}
\label{Tab:results}
\vskip 0.15in
\begin{center}
\begin{scriptsize}
\resizebox{\textwidth}{!}{
\begin{tabular}{llc@{\hspace{0.5em}}ccc@{\hspace{0.7em}}ccc@{\hspace{0.7em}}ccc}
\toprule
\textbf{Dataset} & \textbf{Model} & \textbf{Risk} & \multicolumn{3}{c}{\textbf{TD-AUC}} & \multicolumn{3}{c}{\textbf{TD-CI}} & \multicolumn{3}{c}{\textbf{Brier Score}} \\
\cmidrule(lr){4-6} \cmidrule(lr){7-9} \cmidrule(lr){10-12}
& & & $q_{.25}$ & $q_{.50}$ & $q_{.75}$ & $q_{.25}$ & $q_{.50}$ & $q_{.75}$ & $q_{.25}$ & $q_{.50}$ & $q_{.75}$ \\
\midrule
\multirow{6}{*}{FHS} 
& \multirow{2}{*}{CRISPNAM-FG}  & 1 & \textbf{0.856±0.025} & \textbf{0.832±0.007} & \textbf{0.812±0.010} & \textbf{0.770±0.026} & \textbf{0.755±0.008} & \textbf{0.745±0.007} & 0.122±0.102 & 0.202±0.141 & 0.249±0.153 \\
& & 2 & \textbf{0.812±0.049} & \textbf{0.789±0.016} & \textbf{0.776±0.019} & 0.727±0.054 & 0.716±0.019 & 0.714±0.021 & 0.044±0.011 & 0.092±0.027 & 0.143±0.038 \\
\cmidrule(lr){2-12}

& \multirow{2}{*}{NFG} & 1 & 0.678±.160 & 0.673±.149 & 0.666±.133 & 0.632±.150 & 0.629±.141 & 0.628±.134 & 0.058±.005 & 0.102±.005 & 0.134±.007 \\
& & 2 & 0.617±.126 & 0.629±.117 & 0.620±.136 & 0.612±.156 & 0.611±.148 & 0.610±.152 & 0.041±.003 & 0.077±.006 & 0.113±.009 \\
\cmidrule(lr){2-12}
& \multirow{2}{*}{DEEPHIT}  & 1 & 0.854±.019 & 0.831±.012 & 0.807±.009 & 0.738±.030 & 0.729±.018 & 0.724±.017 & \textbf{0.056±.006} & \textbf{0.101±.004} & \textbf{0.134±.004} \\
& & 2 & 0.796±.053 & 0.779±.028 & 0.776±.031 & 0.737±.030 & 0.728±.018 & 0.724±.017 & \textbf{0.038±.003} & \textbf{0.072±.004} & \textbf{0.109±.005} \\
\midrule
\multirow{6}{*}{SUP} 
& \multirow{2}{*}{CRISPNAM-FG} & 1 & \textbf{0.880±0.031} & \textbf{0.848±0.027} &\textbf{ 0.848±0.031} & \textbf{0.857±0.026} &\textbf{ 0.830±0.017} & \textbf{0.833±0.018} & 0.193±0.120 & 0.307±0.161 & 0.329±0.162 \\
& & 2 & 0.947±0.043 & 0.877±0.091 & 0.832±0.112 & \textbf{0.894±0.007} & \textbf{0.829±0.005 }& \textbf{0.793±0.006} & 0.144±0.035 & 0.251±0.075 & 0.312±0.100 \\
\cmidrule(lr){2-12}

& \multirow{2}{*}{NFG}       & 1 & 0.790±.036 & 0.777±.023 & 0.797±.020 & 0.620±.038 & 0.513±.025 & 0.438±.025 & 0.044±.003 & 0.100±.005 & 0.115±.005 \\
& & 2 & 0.902±.004 & 0.840±.003 & 0.812±.005 & 0.847±.008 & 0.741±.007 & 0.701±.010 & 0.092±.004 & 0.151±.004 & 0.176±.002 \\
\cmidrule(lr){2-12}
& \multirow{2}{*}{DEEPHIT}  & 1 & 0.779±.205 & 0.640±.222 & 0.665±.188 & 0.746±.013 & 0.540±.010 & 0.497±.011 & \textbf{0.039±.003} & \textbf{0.089±.002} & \textbf{0.101±.003} \\
& & 2 & \textbf{0.955±.025} & \textbf{0.893±.069} & \textbf{0.860±.083} & 0.863±.008 & 0.745±.005 & 0.697±.006 & \textbf{0.087±.005 }& \textbf{0.144±.006} & \textbf{0.171±.003} \\
\midrule
\multirow{6}{*}{PBC} 
& \multirow{2}{*}{CRISPNAM-FG} & 1 & 0.992±0.006 & \textbf{0.968±0.023} & 0.956±0.028 &\textbf{ 0.838±0.029} & \textbf{0.813±0.016} & \textbf{0.796±0.015}& 0.199±0.087 & 0.249±0.089 & 0.270±0.080 \\
& & 2 & \textbf{0.980±0.029} & \textbf{0.986±0.012} &\textbf{ 0.988±0.010} & \textbf{0.834±0.029} & \textbf{0.857±0.027} & \textbf{0.868±0.024} & 0.121±0.054 & 0.179±0.072 & 0.190±0.066 \\
\cmidrule(lr){2-12}

& \multirow{2}{*}{NFG}       & 1 & 0.853±.023 & 0.835±.011 & 0.824±.027 & 0.782±.017 & 0.765±.013 & 0.756±.016 & 0.143±.020 & 0.152±.008 & 0.170±.013 \\
& & 2 & 0.491±.055 & 0.491±.055 & 0.537±.056 & 0.227±.022 & 0.238±.011 & 0.245±.016 & 0.092±.054 & 0.135±.073 & 0.164±.082 \\
\cmidrule(lr){2-12}
& \multirow{2}{*}{DEEPHIT}  & 1 & \textbf{0.994±.008} & 0.965±.028 & \textbf{0.959±.034} & 0.822±.016 & 0.796±.019 & 0.776±.013 & \textbf{0.100±.009} & \textbf{0.130±.004} &\textbf{ 0.152±.011} \\
& & 2 & 0.978±.036 & 0.983±.027 & 0.987±.020 & 0.728±.024 & 0.705±.019 & 0.690±.016 & \textbf{0.037±.006} & \textbf{0.056±.003} & \textbf{0.065±.004} \\


\bottomrule
\end{tabular}
}
\end{scriptsize}
\end{center}
\vskip 0.1in
\footnotesize
\textbf{Notes:} Dataset = (FHS: Framingham Heart Study, SUP: SUPPORT2, PBC: Primary Biliary Cirrhosis); 
Model= (CRISPNAM-FG: CRISPNAM -Fine Gray, NFG: Neural Fine Gray, DEEPHIT: DeepHit); 
Risk =  (1: Primary, 2: Competing)
\end{table}

CRISPNAM-FG demonstrates superior discriminative performance, consistently achieving the highest TD-AUC and TD-CI values across most dataset-risk combinations, particularly excelling in the FHS dataset where it outperforms alternatives across all quantiles for both discrimination metrics. However, this enhanced discriminative ability comes at the cost of calibration performance, as evidenced by systematically higher Brier scores compared to DEEPHIT. DEEPHIT exhibits a contrasting profile, consistently producing the lowest Brier scores across all datasets and risks, indicating superior probabilistic calibration, while maintaining competitive but generally lower discriminative performance than CRISPNAM-FG.

\subsection{GEMINI Dataset Results}

\begin{table}[ht!]
\caption{Comparative 5-fold performance metrics between CRISPNAM-FG, DEEPHIT and  Neural Fine Gray models for the GEMINI dataset.}
\label{Tab:results-gemini}
\vskip 0.15in
\begin{center}
\begin{scriptsize}
\resizebox{\textwidth}{!}{
\begin{tabular}{llc@{\hspace{0.5em}}ccc@{\hspace{0.7em}}ccc@{\hspace{0.7em}}ccc}
\toprule
\textbf{Dataset} & \textbf{Model} & \textbf{Risk} & \multicolumn{3}{c}{\textbf{TD-AUC}} & \multicolumn{3}{c}{\textbf{TD-CI}} & \multicolumn{3}{c}{\textbf{Brier Score}} \\
\cmidrule(lr){4-6} \cmidrule(lr){7-9} \cmidrule(lr){10-12}
& & & $q_{.25}$ & $q_{.50}$ & $q_{.75}$ & $q_{.25}$ & $q_{.50}$ & $q_{.75}$ & $q_{.25}$ & $q_{.50}$ & $q_{.75}$ \\
\midrule
\multirow{6}{*}{GEMINI} & \multirow{2}{*}{CRISPNAM-FG} & 1 & \textbf{0.766±.042} & \textbf{0.763±.047} & \textbf{0.763±.052} & \textbf{0.711±.018} & 0.702±.024 & \textbf{0.702±.022} & 0.013±.002 & 0.024±.004 &\textbf{ 0.033±.007} \\
& & 2 & \textbf{0.739±.057} & \textbf{0.726±.058} & \textbf{0.707±.060} & 0.720±.004 & 0.707±.005 & \textbf{0.690±.011} & 0.280±.186 & 0.372±.218 & 0.402±.212 \\
\cmidrule(lr){2-12}
& \multirow{2}{*}{NFG} & 1 & 0.518±.033 & 0.524±.035 & 0.508±.049 & 0.511±.032 & 0.515±.033 & 0.500±.044 & 0.067±.004 & 0.111±.004 & 0.154±.007 \\
& & 2 & 0.440±.005 & 0.396±.012 & 0.367±.025 & 0.550±.014 & 0.532±.015 & 0.529±.020 & 0.016±.002 & \textbf{0.034±.004} &\textbf{ 0.054±.005} \\
\cmidrule(lr){2-12}
& \multirow{2}{*}{DEEPHIT} & 1 & 0.697±.010 & 0.689±.014 & 0.684±.019 & 0.709±.011 & 0.704±.006 & 0.698±.009 &\textbf{ 0.012±.001} & \textbf{0.019±.002} & 0.089±.130 \\
& & 2 & 0.704±.006 & 0.671±.008 & 0.655±.015 & \textbf{0.748±.008} & \textbf{0.720±.010} & 0.690±.022 &\textbf{ 0.052±.002} & 0.077±.003 & 0.142±.098 \\
\bottomrule
\end{tabular}
}
\end{scriptsize}
\end{center}
\end{table}

As shown in Table~\ref{Tab:results-gemini}, for the GEMINI dataset, the CRISPNAM-FG model shows impressive results for both discriminatory  metrics (TD-AUC, TD-CI). While the Brier score for the competing risk is poorer compared to other models, its calibration metric is very good for the primary risk of interest. 

\begin{figure}[h!]
    \centering
    \includegraphics[width=0.70\linewidth]{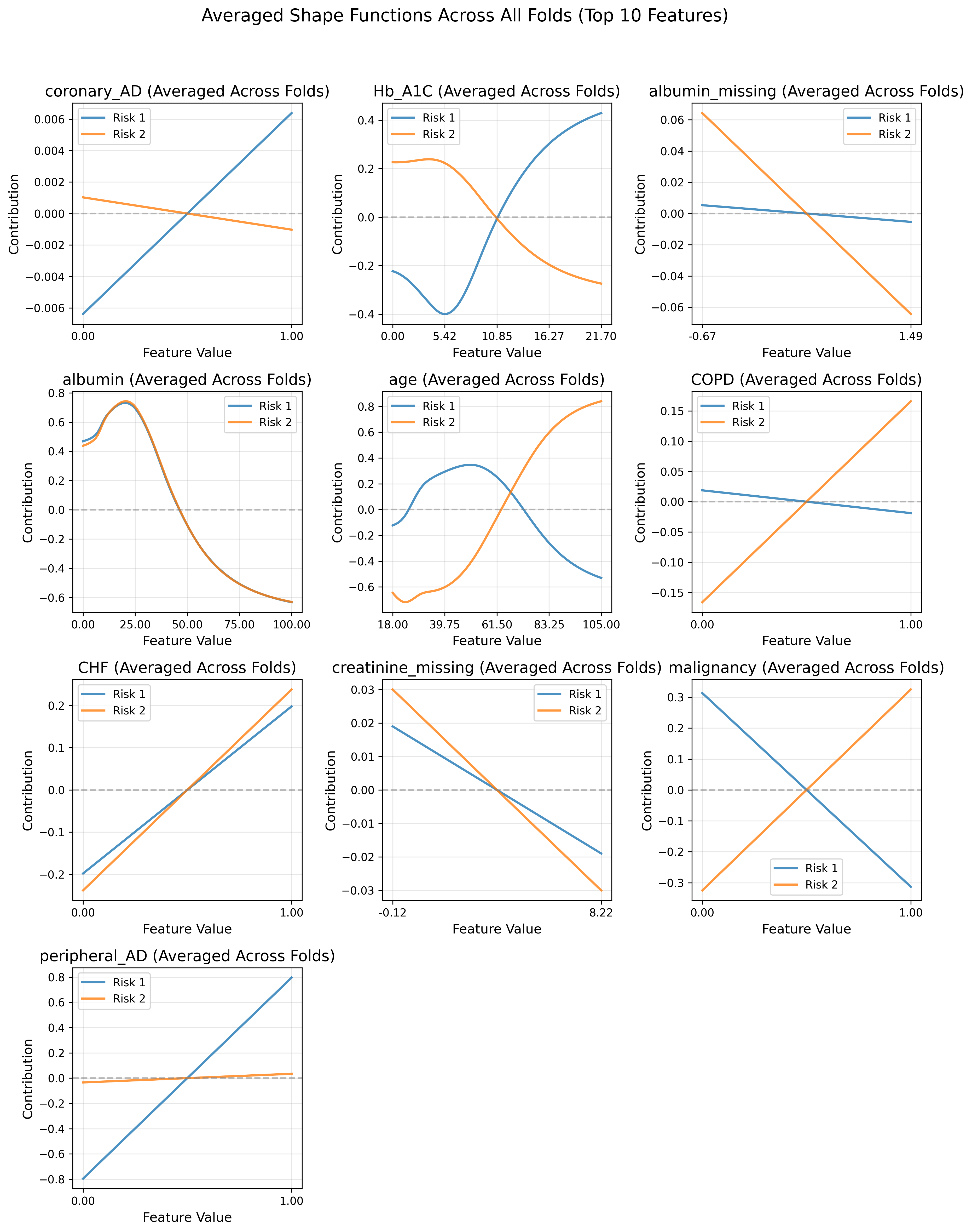}
    \caption{Shape functions generated by CRISPNAM-FG for the GEMINI Foot Complication dataset, averaged across folds. Ten features per risk are shown, randomly selected from the 20 most important.}
    \label{fig:shape_gemini}
\end{figure}

\begin{figure}[ht!]
    \centering
    \begin{subfigure}{0.45\textwidth}
        \centering
        \includegraphics[width=0.85\textwidth]{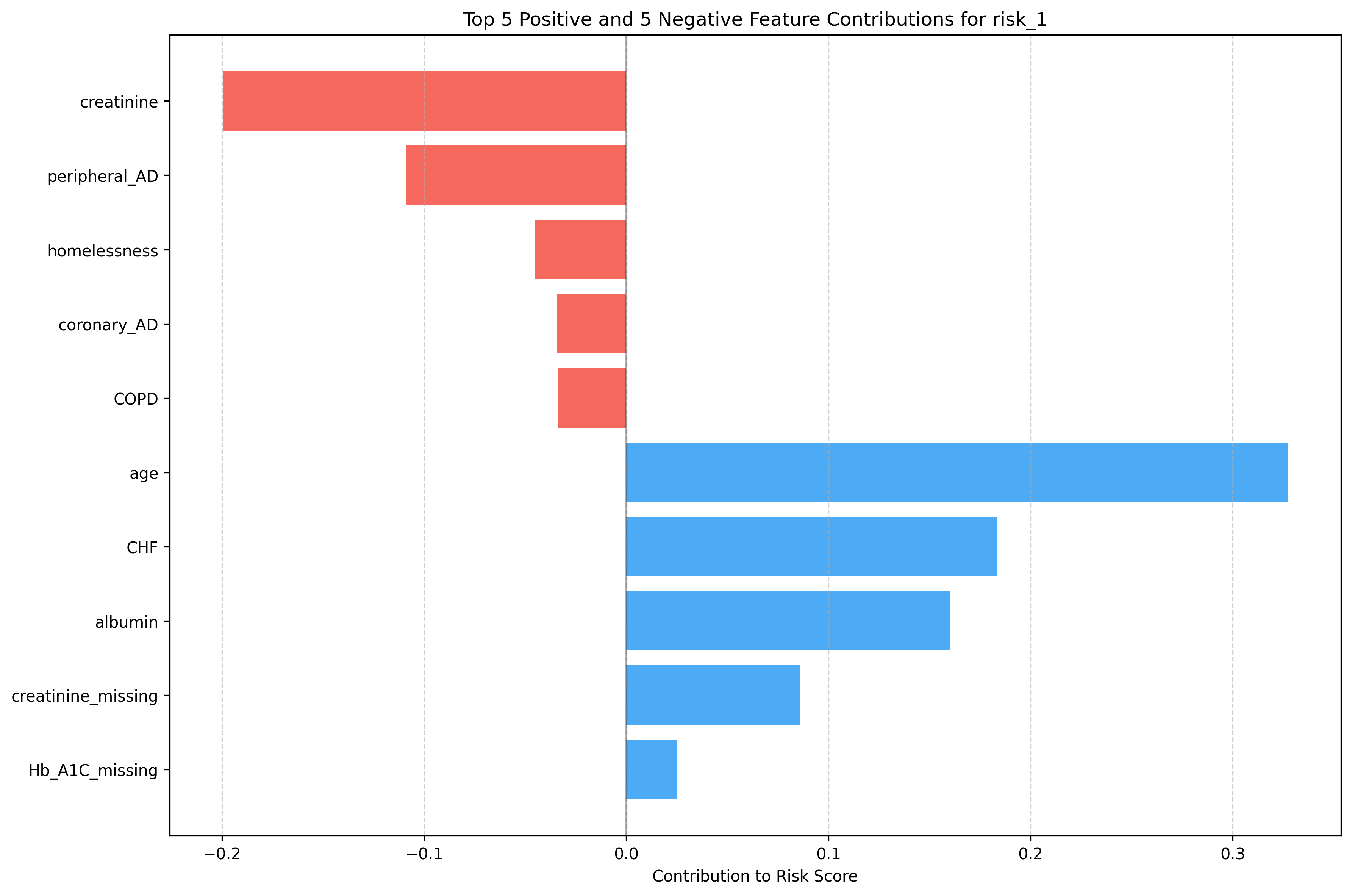}
        \caption{Risk of Future Foot Complication (Risk 1)}
        \label{fig:sub1}
    \end{subfigure}
    \hfill
    \begin{subfigure}{0.45\textwidth}
        \centering
        \includegraphics[width=0.85\textwidth]{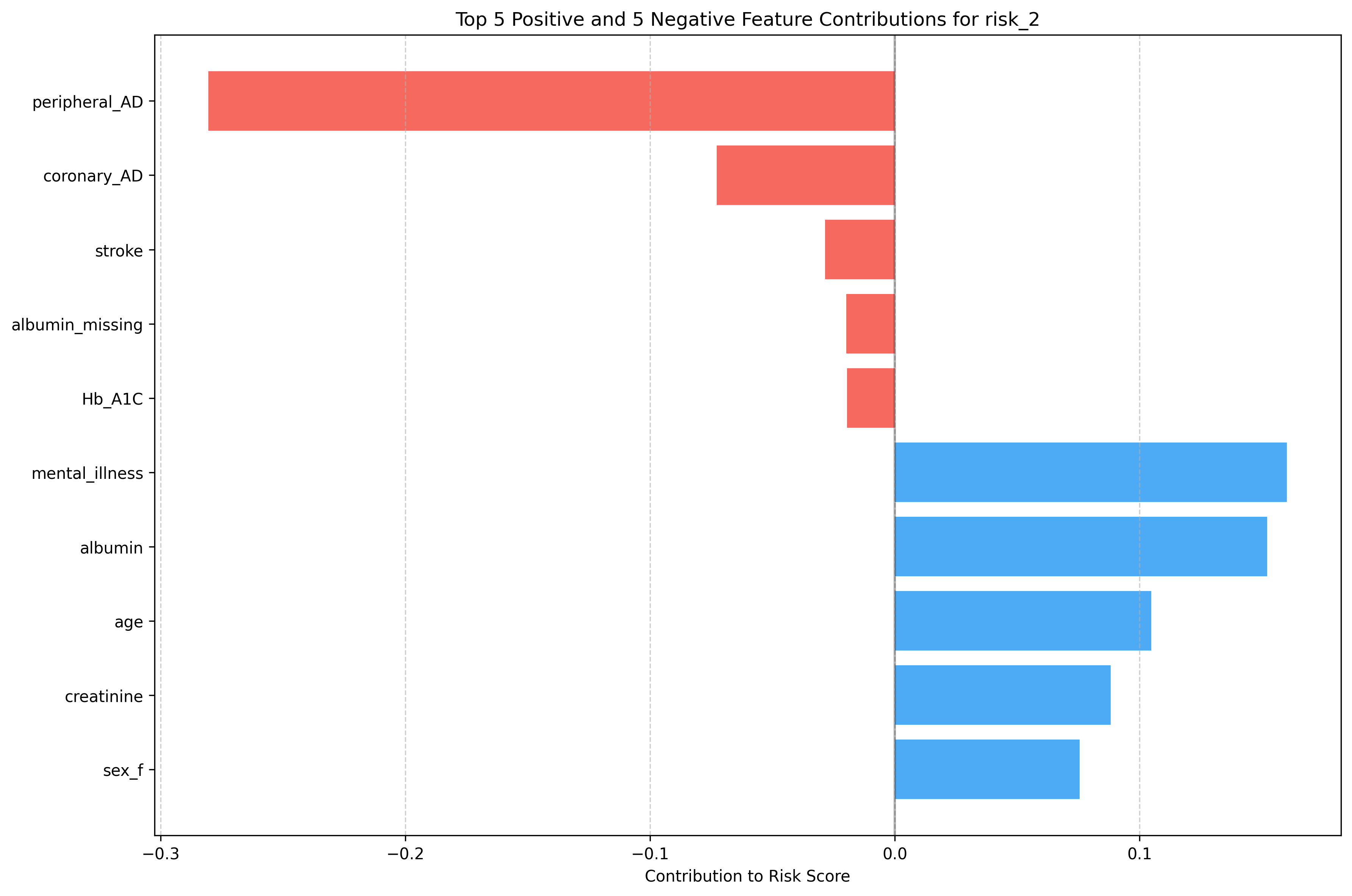}
        \caption{Risk of Death (Risk 2)}
        \label{fig:sub2}
    \end{subfigure}
    \caption{Feature Importances computed by CRISPNAM-FG for Competing Risks for the GEMINI Dataset}
    \label{fig:main-gemini-feat}
\end{figure}

Figure~\ref{fig:shape_gemini} illustrates the shape functions for 10 significant features contributing to model predictions. The presence of comorbidities, specifically coronary artery disease (\texttt{coronary\_ad}) and peripheral artery disease (\texttt{peripheral\_ad}), shows increased contribution to the risk of future foot complications, consistent with established medical knowledge. Similarly, elevated HbA1c levels, which is a marker of poor glycemic control, demonstrate increased contribution to foot complication risk, aligning with the known pathophysiology of diabetic foot disease.
Age exhibits distinct patterns for the two outcomes: the risk contribution for foot complications increases until approximately 60 years before declining, whereas mortality risk increases monotonically with age. This divergence after age 60 likely reflects competing risks, where older patients succumb to systemic complications before developing foot-specific morbidity.
The model correctly identifies that chronic obstructive pulmonary disease (\texttt{COPD}) does not contribute substantially to foot complication risk but significantly increases mortality risk. Similarly, malignancy shows stronger association with mortality than with foot complications. Finally, missing values for creatinine and albumin may indicate selective test ordering, where clinicians did not order these tests in the absence of clinical suspicion for renal dysfunction or hypoalbuminemia-related complications.

\clearpage

\section{Conclusions}
We have proposed an intrinsically interpretable model called CRISPNAM-FG which uses the Fine-Gray formulation to predict the Cumulative Incidence Function for competing risks scenarios in survival analysis. Our benchmarking results indicate that our model produces very competitive discrimination scores (TD-AUC and TD-CI respectively) while the calibration performance of our model lags behind that of models such as DEEPHIT. Additionally, a novel aspect of our work is the application of CRISPNAM-FG to a real world dataset to predict future foot complications post-discharge for a large cohort of diabetic patients in Ontario. On this dataset, our model's performance was better than competing approaches. The core value add for mission critical applications in healthcare is the inherent interpretability provided by our model. The analysis of the shape function plots show in detail how each feature contributes to each risk, which can be valuable for trust and model validation.
Since the goal is to prioritize interpretability, our  model does not capture feature level interactions that may add complexity to the model and limits visualization to up to pairwise interactions only. Future improvements for the model would include investigating temporal feature nets which can capture temporal variation in feature contributions for each risk and developing an improved loss function that could potentially improve model's calibration performance.

\section*{Research Ethics}

Research ethics approval for model development has been obtained from all participating GEMINI
hospitals, as per the study protocols approved by Unity Health Toronto's REB (SMH REB: 20-216, CTO
ID: 3344) and in full compliance with PHIPA and TCPS2.

\clearpage
\bibliography{references}
\bibliographystyle{plainnat}

\end{document}